\documentclass{article}

\usepackage[T1]{fontenc}
\usepackage{iclr2027_conference,times}

\usepackage{amsmath,amsfonts,bm}

\def\eqref#1{equation~\ref{#1}}

\def\1{\bm{1}}

\DeclareMathAlphabet{\mathsfit}{\encodingdefault}{\sfdefault}{m}{sl}
\SetMathAlphabet{\mathsfit}{bold}{\encodingdefault}{\sfdefault}{bx}{n}

\usepackage{hyperref}
\hypersetup{hidelinks}
\usepackage{url}
\usepackage{booktabs}
\usepackage{multirow}
\usepackage{colortbl}
\usepackage{graphicx}
\usepackage{capt-of}
\usepackage{wrapfig}
\usepackage{amssymb}
\usepackage{amsmath}
\usepackage{xcolor}
\usepackage{tikz}
\usepackage{pgfplots}

\definecolor{citeblue}{RGB}{52,101,164}

\hypersetup{
    colorlinks=true,
    citecolor=citeblue,
    linkcolor=red,
    urlcolor=blue
}

\pgfplotsset{compat=1.18}

\usetikzlibrary{ arrows.meta, positioning, fit, backgrounds, shapes.geometric }

\title{IRSTD-Agent: Agentic Infrared Small Target Detection via Zoom-Guided Interaction Learning}

\author{
Jiawen Xi$^{1}$ \quad
Yu Zhang$^{1}$ \quad
Tianyi Zhao$^{1}$ \quad
Zhu Liu$^{2}$ \quad
Maoxun Yuan$^{1,*}$ \quad
Xingxing Wei$^{1,*}$ \\
$^{1}$Beihang University \qquad
$^{2}$Dalian University of Technology \quad
$^{*}$Corresponding authors
}

\iclrfinalcopy

\begin{document}

\maketitle


\begin{abstract}

Infrared small-target detection plays an important role in maritime monitoring and aerial surveillance. Although multimodal large language models (MLLMs) offer promising capabilities for visual understanding, existing MLLM-based approaches struggle to precisely localize infrared small targets. In this paper, we propose \textbf{IRSTD-Agent}, an agentic framework for infrared small target detection through dynamic visual search. The framework enables an MLLM to adaptively determine where and at what scale to inspect an image and progressively gather fine-grained visual evidence for precise target localization. Five complementary visual tools (PROPOSAL, ZOOM, DETECT, DROP and REFINE) support object candidate discovery, adaptive observation, target localization, hypothesis rejection, and target extent refinement, together enabling a coordinated search process over original-resolution images. To teach the MLLMs to conduct this search, we introduce Zoom-guided Interaction Learning, which uses annotation-derived interaction trajectories to supervise tool selection and the corresponding arguments. Through extensive experiments on WideIRSTD-Full and IRSTD-1k datasets, we demonstrate that IRSTD-Agent outperforms the evaluated vision-language models and enhances the precise localization capabilities of MLLMs in IRSTD tasks.
\end{abstract}

\section{Introduction}

Infrared small target detection (IRSTD) aims to identify and accurately localize dim thermal targets in cluttered backgrounds, supporting applications such as maritime monitoring and aerial surveillance. Although specialized deep learning-based approaches have made substantial progress through feature enhancement, context aggregation, and background suppression~\citep{li2023dnanet,liu2024mshnet,yuan2025nsfpn}, they are typically developed as task-specific models, thereby limiting their generalization capabilities in visual understanding. With the development of multimodal large language models, it has become possible to possess capabilities similar to human visual intelligence and to handle complex visual understanding tasks~\citep{li2025dyfo,yuan2025unirgb}. 

\begin{figure}[!t]
    \centering
    \includegraphics[width=1\linewidth]{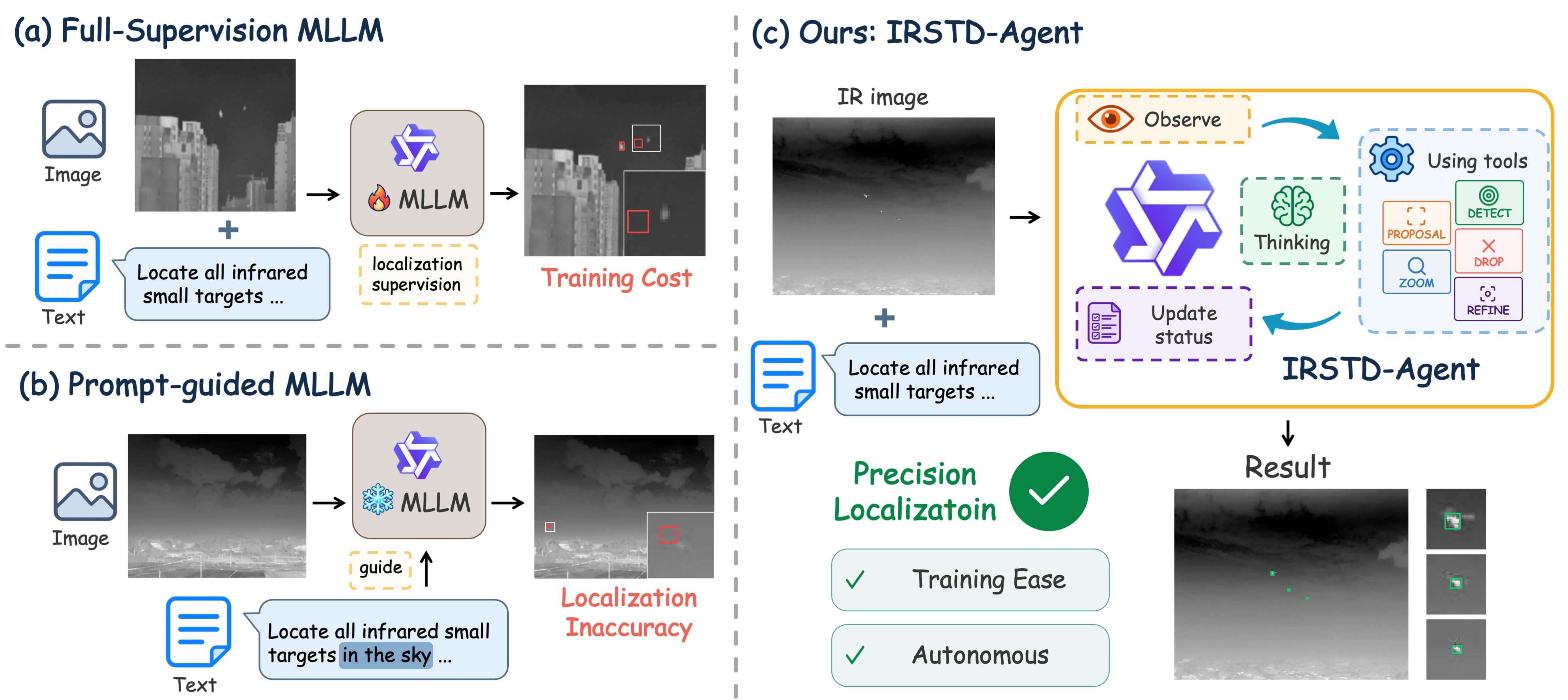}
    \caption{Motivation for IRSTD-Agent. (a) Full-supervision MLLMs adapt to infrared target localization through task-specific supervision, incurring annotation and training costs. (b) Prompt-guided MLLMs use auxiliary language cues without updating model parameters, but fail to precisely localize small infrared targets. (c) Our IRSTD-Agent employs dynamic visual search to acquire local evidence for target recognition and precise localization.}
    \label{fig:intro_motivation}
\end{figure}

Existing efforts to improve the capability of MLLMs in IRSTD task can be broadly grouped into two categories: infrared-domain fine-tuning and cue-guided prompting. As shown in Figure~\ref{fig:intro_motivation}(a), the first strategy fine-tunes MLLMs on infrared image–text pairs and task-specific instructions to improve their understanding of infrared imagery, such as Infrared-LLaVA and IRGPT~\citep{jiang2024infrared,cao2025irgpt}. However, they require substantial annotation effort and incur considerable computational overhead for model finetuning. More importantly, many state-of-the-art MLLMs~\citep{achiam2023gpt,team2023gemini} are proprietary and cannot provide access to model parameters, making finetuning inapplicable to some of the strongest existing models and limiting its scalability to rapidly evolving MLLMs. Another strategy introduces auxiliary language cues to guide infrared small target localization, as shown in Figure~\ref{fig:intro_motivation}(b). For example, MIRSAM~\citep{zhang2025mirsam} incorporates textual descriptions of target location and appearance through CLIP-based prompts to support infrared small target segmentation. Similarly, LGNet~\citep{chib2026leveraging} leverages MLLMs to generate descriptions of infrared small target locations and incorporates these descriptions into a dedicated detection network to improve performance. 
\textbf{Although these strategies offer promising directions for adapting MLLMs to IRSTD without parameter updates, the effectiveness of precisely localizing infrared small targets remains to be established}.

To address these challenges, we propose an \textbf{IRSTD-Agent}, a framework that formulates IRSTD as a dynamic visual search process, 
as illustrated in Figure~\ref{fig:intro_motivation}(c). Instead of relying on a single fixed observation,  our approach enables an MLLM to actively acquire local visual evidence and adapt its search through sequential tool interactions. Given an infrared image, the agent identifies candidate regions and adaptively determines where and at what scale to inspect, gathering evidence to localize targets and reject background clutter. To support this process, we introduce five complementary visual tools: \textbf{PROPOSAL} for candidate discovery, \textbf{ZOOM} for adaptive observation, \textbf{DETECT} for target localization, \textbf{DROP} for clutter rejection, and \textbf{REFINE} for target extent refinement. By coordinating these tools, our IRSTD-Agent systematically examines object candidate regions and extracts fine-grained visual evidence from original-resolution images, thereby achieving the recognition and precise localization of  infrared small targets through an adaptive search process. To teach the MLLM to guide this search, we introduce \textbf{Z}oom-guided \textbf{I}nteraction \textbf{L}earning (\textbf{ZIL}), which uses annotation-derived interaction trajectories to supervise tool selection and the corresponding arguments. Through ZIL, the MLLM learns to select informative views, localize targets, and reject distractors based on the acquired visual evidence. Our contributions are summarized as follows:

\begin{itemize}
    \item We introduce IRSTD-Agent, the first agentic framework for infrared small target detection, enabling the model to actively acquire local observations and use visual tools for adaptive visual search and precise target localization in cluttered infrared scenes.

    \item We design five active visual tools (PROPOSAL, ZOOM, DETECT, DROP and REFINE) for IRSTD-Agent to perform dynamic visual search and introduce zoom-guided interaction learning to teach MLLMs to select appropriate search tools and their parameters based on visual evidence, using annotation-derived interaction trajectories as supervision.

    \item Extensive experiments conducted on the WideIRSTD-Full and IRSTD-1K datasets demonstrate that IRSTD-Agent consistently outperforms existing MLLM-based methods and accurately localizes infrared small targets across various infrared image scenarios.
\end{itemize}

\section{Related Work}
\paragraph{Infrared Small-Target Detection.}
Modern IRSTD methods develop specialized architectures and supervision for weak-target representation and localization. DNANet~\citep{li2023dnanet} strengthens hierarchical feature propagation, MSHNet~\citep{liu2024mshnet} introduces scale- and location-sensitive supervision, and NS-FPN~\citep{yuan2025nsfpn} improves multi-scale feature interaction through frequency-guided feature fusion. Other studies explore unified encoder--decoder architectures~\citep{wu2023uiunet}, target-adapted convolutions~\citep{yang2025pconv}, and learning from sparse point annotations~\citep{yu2025pal}. AuxDet~\citep{shi2026auxdet} further studies omni-domain IRSTD by incorporating auxiliary acquisition metadata.Recent work has also introduced vision-language priors into IRSTD. SAIST~\citep{zhang2025saist} integrates textual guidance with SAM for infrared small-target segmentation. Overall, these approaches form specialized solutions tailored to infrared small-target detection and are typically deployed through predefined detection pipelines.

\paragraph{Vision-Language Models for Visual Localization.}
Recent VLMs have extended multimodal understanding to spatial localization and detection. General-purpose MLLMs such as Qwen3.5/3.8~\citep{qwen2026qwen35,qwen2026qwen38} and InternVL3.5~\citep{wang2025internvl35} can follow visual localization instructions and produce coordinate-based predictions. Earlier grounding models explicitly align natural-language expressions with image regions. MDETR~\citep{kamath2021mdetr} performs end-to-end text-conditioned object detection through joint vision-language reasoning, while GLIP~\citep{li2022glip} unifies object detection and phrase grounding within a grounded language-image pre-training framework.Models such as LocateAnything~\citep{nvidia2026locateanything}, SenseNova-Vision~\citep{sensenova2026vision}, Grounding DINO~\citep{liu2024groundingdino}, and Florence-2~\citep{xiao2024florence2} further develop language-conditioned grounding and structured spatial prediction.

\paragraph{Visual Chain-of-Thought and Structured Perception Pipelines.}
Recent work incorporates visual evidence acquisition into multimodal reasoning along two main directions. One line extends chain-of-thought reasoning with intermediate visual interactions, allowing models to identify, revisit, or zoom into relevant regions during reasoning in a more deliberate and iterative manner. Representative approaches include V*, Visual CoT, DeepEyes, and Pixel Reasoner~\citep{wu2024vstar,shao2024visualcot,zheng2026deepeyes,wang2025pixelreasoner}, which study such capabilities across grounding, fine-grained perception, and multimodal reasoning tasks. Another line organizes visual processing through structured workflows. ZoomEye performs hierarchical image exploration through tree search~\citep{shen2025zoomeye}, ZoomEarth develops an adaptive cropping framework for high-resolution remote-sensing tasks~\citep{liu2026zoomearth}, and AgentDet coordinates specialized roles through a shared-blackboard detection pipeline~\citep{li2026agentdet}.

\section{IRSTD-Agent}
\label{sec:method}

\begin{figure}[t]
    \centering
    \includegraphics[width=\textwidth]{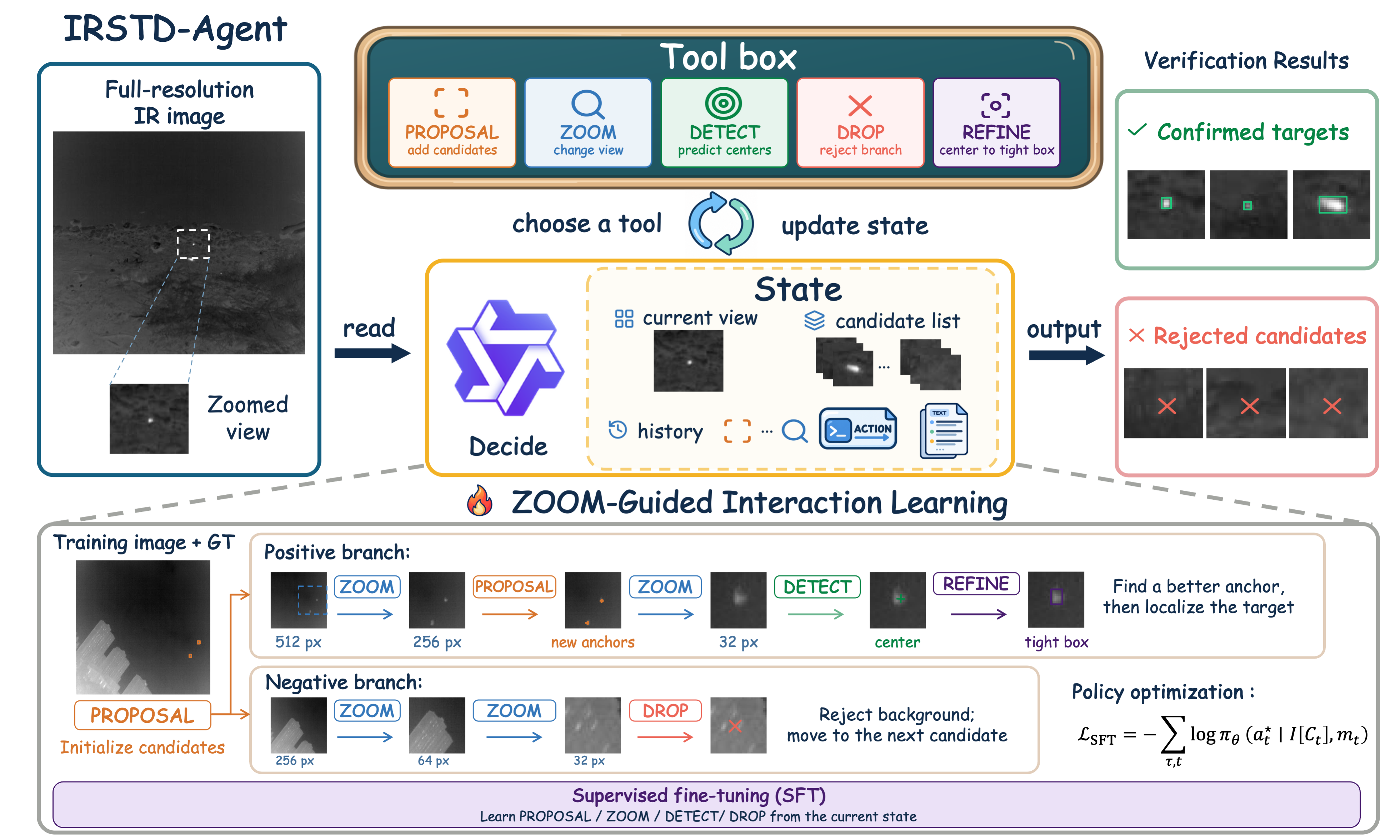}
    \caption{Overview of IRSTD-Agent. The policy interacts with an infrared image through visual tools, acquiring local observations and accumulating target localizations in an evidence workspace. Annotation-derived interaction trajectories supervise tool selection and arguments. Candidate discovery and appearance-guided refinement support the search and localization process.}
    \label{fig:irstd_agent_overview}
\end{figure}

IRSTD-Agent formulates infrared small-target detection as a dynamic visual search process over the original-resolution image. Rather than predicting all targets from a single resized observation, the agent repeatedly selects visual tools to discover candidate regions, acquire higher-resolution evidence, localize visible targets, and reject unsupported hypotheses. Figure~\ref{fig:irstd_agent_overview} summarizes this process: the current view, candidate list, and interaction history inform each tool call, whose execution updates the state for subsequent decisions. Target localizations are accumulated across search branches, while unsupported candidates are rejected. We first describe how the agent selects and executes these tools, and then detail each tool and the annotation-derived interaction learning.

\subsection{Agent Formulation}
\label{sec:agent_design}

Figure~\ref{fig:irstd_agent_overview} summarizes the framework, linking tool-based visual search and the evidence workspace with annotation-derived interaction supervision.

Given an infrared image $I\in\mathbb{R}^{H\times W}$ , IRSTD-Agent performs detection through successive visual-tool interactions. The agent is equipped with five tools,
\begin{equation}
    \mathcal U=
    \{\mathtt{PROPOSAL},\mathtt{ZOOM},
      \mathtt{DETECT},\mathtt{DROP},\mathtt{REFINE}\}.
    \label{eq:tool_space}
\end{equation}
These tools serve different stages of visual search: \textsc{Proposal} discovers candidate locations, \textsc{Zoom} acquires a new local observation, \textsc{Detect} records visible targets, \textsc{Drop} rejects unsupported hypotheses, and \textsc{Refine} estimates the spatial extent of an existing localization.

At interaction step $t$, the agent observes the current view $I[C_t]$, where $C_t=(X_t,Y_t,W_t)$ specifies the view center $(X_t,Y_t)$ and window size $W_t$ in the original image. The interaction state $S_t$ maintains the information required to continue the search, including the current view, candidate hypotheses, interaction history, accumulated detections, and remaining budget. The model-visible context $m_t$ is obtained by serializing the information in $S_t$ that is exposed to the policy.

Conditioned on the current view and  context, the search policy predicts a structured tool call
\begin{equation}
    a_t
    \sim
    \pi_\theta(\cdot\mid I[C_t],m_t),
    \label{eq:agent_policy}
\end{equation}
where $a_t$ specifies the selected search tool together with its required arguments. The search policy selects from
$\mathcal U_{\mathrm{search}}
=\{\mathtt{PROPOSAL},\mathtt{ZOOM},\mathtt{DETECT},\mathtt{DROP}\}$.
\textsc{Refine} operates on an existing detection and is handled separately.

Once a tool is selected, it is executed on the original image and updates the interaction state:
\begin{equation}
    S_{t+1}
    =
    \mathcal E(I,S_t,a_t),
    \label{eq:state_transition}
\end{equation}
where $\mathcal E$ denotes the tool executor. The updated state is then serialized into the context $m_{t+1}$, together with the latest interaction information, for the next agent decision.

\subsection{Active Visual Tools}
\label{sec:verification_agent}
\label{sec:execution}

\begin{figure}[t]
    \centering
    \includegraphics[width=\linewidth]{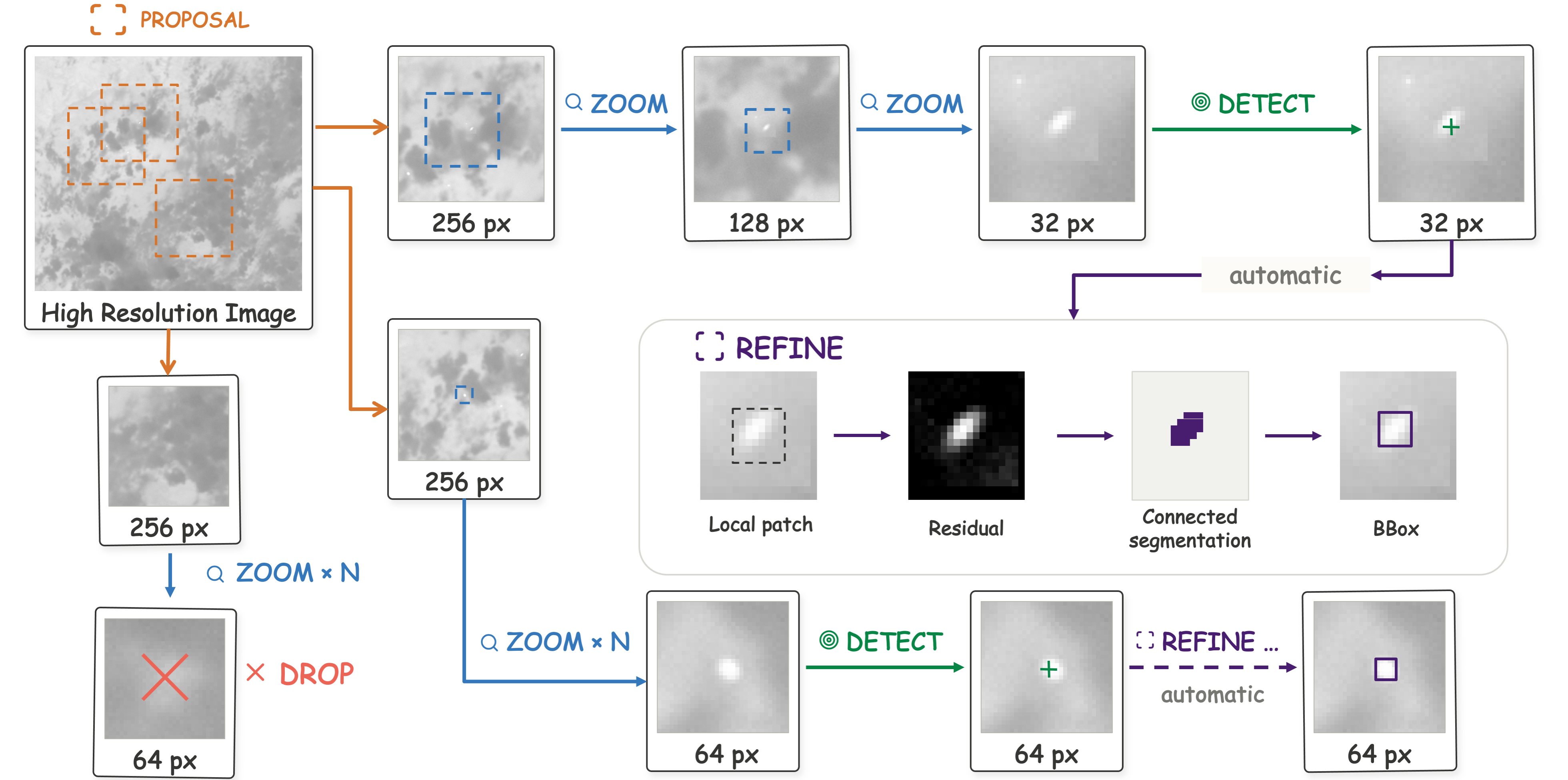}
    \caption{Visual-tool interaction in IRSTD-Agent.
The agent iteratively selects tools based on the current view and accumulated interaction context, while tool execution updates the search process and acquired evidence. Detection and refinement produce the final target localizations and extents.}
    \label{fig:tool_workflow}
\end{figure}

Each visual tool operates directly on native-resolution pixels from the original image $I$. Given the current view $C_t$, the tools either update the view, introduce candidate hypotheses, produce target localizations, or refine existing detections, as detailed below. Figure~\ref{fig:tool_workflow} illustrates how successive tool calls update the observations and accumulated evidence to produce target localizations and refined extents.

\paragraph{\textsc{Proposal}: candidate discovery.}
\textsc{Proposal} discovers potential target locations within the current search region during the active visual search process. Given the native-resolution crop $I[C_t]$, it returns a ranked set of candidate hypotheses:
\begin{equation}
    \mathcal Q_t
    =
    \mathtt{PROPOSAL}\!\left(I[C_t]\right)
    =
    \bigl((q_{tk},s_{tk})\bigr)_{k=1}^{K_t},
    \qquad
    s_{t1}\geq\cdots\geq s_{tK_t}.
    \label{eq:proposal_queue}
\end{equation}
Here $q_{tk}$ denotes the $k$-th candidate location mapped to original-image coordinates, and $s_{tk}$ is its proposal score. The returned candidates are search hypotheses rather than final detections. Internally, the tool combines complementary spatial and frequency responses. The spatial branch aggregates multi-scale morphological top-hat filtering, difference-of-Gaussians, local standardized contrast, and patch contrast, while the frequency branch uses FFT-based spectral-residual saliency. Their normalized responses are fused as
\begin{equation}
    R_t=\lambda R_{\mathrm{sp},t}
        +(1-\lambda)R_{\mathrm{fft},t}.
    \label{eq:rank_fusion}
\end{equation}
Candidate peaks are then ranked and spatially suppressed to form $\mathcal Q_t$.

\paragraph{\textsc{Zoom}: evidence acquisition.}
\textsc{Zoom} predicts a new crop-relative center $(x_t,y_t)$ and window size $w_t$ for the next observation during the active visual search process. These parameters are applied to the current view $C_t$ to update the search region:
\begin{equation}
    C_{t+1}
    =
    \mathtt{Zoom}
    \!\left(C_t;\,x_t,y_t,w_t\right).
    \label{eq:zoom}
\end{equation}
Here $(x_t,y_t)$ specifies where to move within the current view, while $w_t$ controls the scale of the next view. The updated region $I[C_{t+1}]$ is then used in the next agent interaction.

\paragraph{\textsc{Detect}: target localization.}
\textsc{Detect} enables IRSTD-Agent itself to localize the targets visible in the current view. Given the current image region $I[C_t]$, the agent predicts a structured set of normalized target centers:
\begin{equation}
    \mathcal P_t
    =
    \mathtt{DETECT}\!\left(I[C_t]\right)
    =
    \{p_{tj}\}_{j=1}^{n_t},
    \label{eq:detect}
\end{equation}
where $p_{tj}$ denotes the normalized center of the $j$-th target predicted by IRSTD-Agent within the current view. The predicted centers are converted to the original-image coordinate system and recorded as image-level detections. The current search branch is then terminated.

\paragraph{\textsc{Drop}: hypothesis rejection.}
\textsc{Drop} is an argument-free action selected when the current visual evidence does not support the presence of a target:
\begin{equation}
    a_t=\mathtt{DROP}
    \quad\Longrightarrow\quad
    \text{terminate the current search branch without producing a detection}.
    \label{eq:drop}
\end{equation}
The rejection is recorded in the interaction history, while no image-level detection is added.

\paragraph{\textsc{Refine}: appearance-based extent refinement.}
Given a detected center $p_i$, \textsc{Refine} uses its location to recover the target extent from the original image. It first initializes a local box around the center, suppresses the local background, and applies adaptive thresholding to isolate target-supporting responses. The connected region associated with the detected center is then extracted, and its spatial extent is used to obtain the refined bounding box. We denote this image-space transformation as
\begin{equation}
    \tilde b_i
    =
    \mathtt{Refine}(I,p_i),
    \label{eq:refine}
\end{equation}
where $\tilde b_i$ denotes the refined bounding box. The foreground pixels of the extracted connected region define a binary mask $M_i$, which is mapped back to the original-image coordinates. Taking the union of these masks produces the image-level segmentation map used for visualization and the evaluation in Appendix~\ref{sec:segmentation_results}. Thus, the mask follows the image-derived target region rather than a filled detection box. If no valid region is obtained, the initialized local box is preserved and no mask is added to the segmentation map.

\subsection{Zoom-Guided Interaction Learning}
\label{sec:policy_learning}
\label{sec:sft_data}

We construct interaction supervision from standard target annotations to teach IRSTD-Agent how to select visual tools and their arguments during search in a supervised manner. Each training trajectory consists of a sequence of agent interactions,
\begin{equation}
    \tau
    =
    \left\{
    \bigl(I[C_t],m_t,a_t^\star\bigr)
    \right\}_{t=0}^{T},
    \label{eq:trajectory}
\end{equation}
where $I[C_t]$ is the current view, $m_t$ is the accumulated interaction context, and $a_t^\star$ is the supervised tool call at step $t$. Successive actions therefore connect what the agent currently observes
with how it should continue the search.

\paragraph{Annotation-derived interaction trajectories.}
The lower panel of Figure~\ref{fig:irstd_agent_overview} illustrates supervision through positive and negative search branches: positive trajectories acquire evidence for target localization, whereas negative trajectories terminate with \textsc{Drop} after inspecting background candidates.

For positive target hypotheses, we construct target-preserving \textsc{Zoom} actions that progressively expose the annotated target at finer spatial scales, followed by \textsc{Detect} supervision once sufficient target evidence is visible. Proposal-offset examples place the target away from the initial candidate location, preventing the agent from simply copying the proposal position and encouraging localization from the visual evidence itself.

Negative hypotheses are supervised with \textsc{Drop} when the inspected region contains background rather than an annotated target. For views containing multiple visible targets, \textsc{Detect} supervision includes all valid targets in current view, allowing a single branch to return multiple detections. \textsc{Proposal} supervision is included in the same interaction to teach candidate discovery during search.

The resulting interactions are organized into complete trajectories $\tau=\{(I[C_t],m_t,a_t^\star)\}_{t=0}^{T}$. After trajectory construction and sampling, the retained trajectories form the supervised fine-tuning
set
\begin{equation}
    \mathcal D_{\mathrm{SFT}}
    =
    \{\tau_i\}_{i=1}^{N}.
    \label{eq:sft_dataset}
\end{equation}
Together, these trajectories teach the agent when to propose, where and how far to zoom, when to detect, and when to reject a hypothesis.

\paragraph{Policy optimization.}
\label{sec:sft_training}
The search policy is optimized over $\mathcal D_{\mathrm{SFT}}$ by
\begin{equation}
    \mathcal L_{\mathrm{SFT}}
    =
    -\sum_{\tau\in\mathcal D_{\mathrm{SFT}}}
     \sum_t
     \log
     \pi_\theta
     \left(
        a_t^\star
        \mid I[C_t],m_t
     \right).
    \label{eq:sft_loss}
\end{equation}
This objective encourages the policy to assign higher probability to the correct tool call $a_t^\star$ given the current view and interaction context, thereby learning both tool selection and its corresponding arguments. Each supervised action is serialized as a structured tool call, while prompt tokens are excluded from the loss during supervised fine-tuning of the policy.

\section{Experiments}
\subsection{Experimental Setup}
\label{sec:evaluation}
\paragraph{Datasets.}
We study two complementary settings. WideIRSTD-Full~\citep{li2025limitirstd,shi2026auxdet} evaluates wide-area target search on its complete 2,000-image test set, preserving each image's native resolution and allowing multiple targets or no targets per image. Its images span a wide range of resolutions and acquisition domains, introducing substantial variation in both scale and scene and making wide-area target search particularly challenging. IRSTD-1k~\citep{zhang2022isnet} is evaluated using its official split of 800 training images and 201 test images at $512\times512$ resolution. Because of overlapping in two datasets, the settings are  evaluated independently.

\paragraph{Implementation Details.}
IRSTD-Agent is built on Qwen3.5-27B, with separate checkpoints independently fine-tuned on WideIRSTD-Full and IRSTD-1k using the zoom-guided SFT. IRSTD-1k uses its independently trained checkpoint, with no weights transferred from WideIRSTD-Full. We apply LoRA with rank 16, $\alpha=32$, and dropout 0.05, while freezing the vision encoder and multimodal projector. Training is performed for one epoch with a learning rate of $1\times10^{-5}$, cosine decay, 5\% warmup, BF16 computation, and gradient checkpointing on NVIDIA A100 80GB GPUs.

\paragraph{Evaluation Metrics.}
MLLM base models are evaluated using their native boxes without additional refinement or NMS. Grounding methods, Visual Reasoning and Active Search methods, and IRSTD-Agent use center-based refinement and box NMS.
We report precision (P), recall (R), and F1-scores for target detection. Following prior IRSTD evaluations~\citep{li2023dnanet,chen2022irstformer}, a prediction is matched to an annotation if their box centers are within 3 pixels in the original image. Matches are established one-to-one using the same criterion for all methods. We additionally report the mean box IoU over matched targets to evaluate localization extent.

\raggedbottom
\subsection{Ablation Studies}
All ablations below cover the official IRSTD-1k datasets. Tool and budget experiments use the main-table Qwen3.5-27B checkpoint. 

\begin{table}[!t]
\centering\scriptsize
\caption{Visual tool ablation on IRSTD-1k datasets. Checkmarks indicate enabled tools. P, R, and F1-scores are reported in percentages. The best results are highlighted in \textbf{bold}.}
\label{tab:ablation}
\setlength{\tabcolsep}{10pt}
\begin{tabular}{cccccrrr}
\toprule
PROPOSAL & ZOOM & DETECT & DROP & REFINE & P & R & F1 \\
\midrule
$\checkmark$ & $\checkmark$ & $\checkmark$ & & & 3.61 & 77.44 & 6.90 \\
$\checkmark$ & & $\checkmark$ & $\checkmark$ & $\checkmark$ & 23.89 & \textbf{86.87} & 37.47\\
$\checkmark$ & $\checkmark$ & $\checkmark$ & $\checkmark$ & & 63.78 & 69.36 & 66.45\\
\rowcolor{gray!8}
$\checkmark$ &$\checkmark$ & $\checkmark$ & $\checkmark$ & $\checkmark$ & \textbf{81.09} & 75.08 & \textbf{77.97}\\
\bottomrule
\end{tabular}
\end{table}

\noindent\textbf{Visual tools.}
Table~\ref{tab:ablation} shows that the tools play complementary roles in the search process. Without DROP, the agent retains many candidate hypotheses, resulting in high recall (77.44\%) but very low precision (3.61\%).Introducing DROP substantially suppresses false detections, increasing precision to 63.78\% and F1-scores to 66.45\%. Replacing adaptive ZOOM with a fixed-scale observation strategy yields higher recall (86.87\%) but much lower precision (23.89\%) and F1-scores (37.47\%), indicating that adaptive evidence acquisition is important for reliable target verification. Finally, REFINE further improves the precision--recall balance, raising F1-scores from 66.45\% to 77.97\%.

\par
\begin{wrapfigure}{r}{0.43\textwidth}
\vspace{-24pt}
\centering\scriptsize
\captionof{table}{Ablation of zoom-guided interaction learning on IRSTD-1k datasets, P/R/F1-scores are percentages.}
\label{tab:training_ablation}
\setlength{\tabcolsep}{8pt}
\begin{tabular}{lrrr}
\toprule
Training & P & R & F1\\
\midrule
w/o SFT & 5.80 & 78.11 & 10.80\\
w/o negative branch & 6.72 & \textbf{78.79} & 12.38\\
\rowcolor{gray!8} Ours & \textbf{81.09} & 75.08 & \textbf{77.97}\\
\bottomrule
\end{tabular}
\par\vspace{6pt}
\centering\scriptsize
\includegraphics[width=\linewidth]{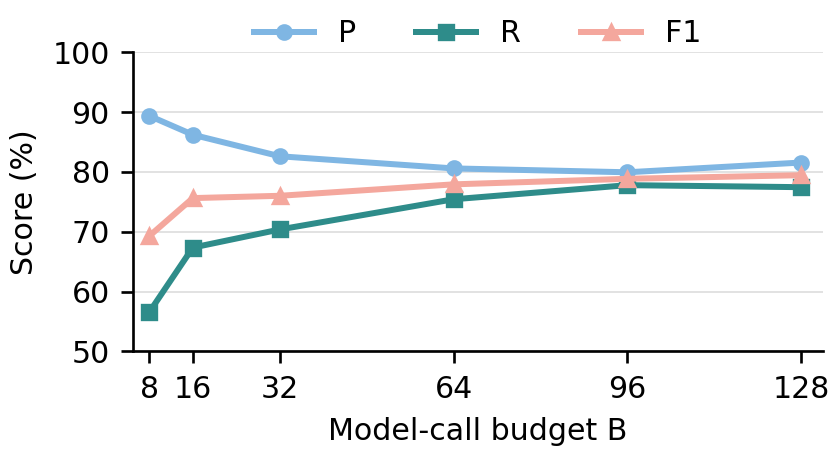}\par
\captionof{figure}{Inference-budget ablation on IRSTD-1k with $B=8,16,32,64,96,128$, varying the maximum model-call budget.}
\label{fig:budget}

\end{wrapfigure}

\par\medskip
\noindent\textbf{Zoom-guided Interaction learning.}
Table~\ref{tab:training_ablation} compares supervision settings and reports the full system as a reference. Under the same native tool-call execution protocol, the model without SFT (w/o SFT) attains 5.80 precision, 78.11 recall, and 10.80 F1-scores; SFT without the negative branch (w/o negative branch) yields 6.72 precision, 78.79 recall, and 12.38 F1-scores. Positive-only supervision thus provides limited improvement in this setting. The full zoom-guided SFT achieves 81.09 precision and 77.97 F1-scores, delivering the best performance.

\par\medskip
\noindent\textbf{Action budget.}
Figure~\ref{fig:budget} varies both model-call  caps over B=\{8,16,32,64,96,128\}, with the checkpoint and proposal/detection caps fixed. F1-scores increases from 69.28 to 79.45 and recall from 56.57 to 77.44. However, F1-scores rises by 0.87 points from 64 to 96 despite a substantial increase in inference time,and by 0.61 points from 96 to 128,indicating diminishing returns.These results indicate that $B=64$ provides a favorable trade-off between detection performance and action budget.

\subsection{ Comparison with SOTA Methods}
\paragraph{Baseline Setup. }
\textit{Open-Source MLLMs} are evaluated by directly prompting them to detect infrared targets from the full image and return bounding boxes. For models with their own coordinate conventions, the outputs are normalized to a common image-coordinate format before evaluation; for example, Qwen predictions follow its native 0--1000 coordinate protocol and are mapped back to the original image size.
\textit{VLM-Based Grounding Models} support multi-object localization or open-vocabulary detection, so they are directly queried with an infrared-target prompt and evaluated on all returned target boxes in the full image.
\textit{Visual Reasoning and Active Search Methods} are adapted to IRSTD through method-specific search, cropping, or reasoning interfaces. In particular, ZoomEye uses Qwen3.5-27B for hierarchical visual search before producing detections. Other methods receive infrared-target queries and structured detection outputs; Appendix~\ref{app:implementation_details} details their training, inference, and output-processing differences.

\begin{table*}[t]
\centering
\caption{Detection results on WideIRSTD-Full and IRSTD-1k. The shared Method and Venue columns are followed by dataset-specific precision, recall, F1-scores, and IoU; values are percentages.}
\label{tab:dataset_comparison}
\small
\setlength{\tabcolsep}{2.5pt}
\renewcommand{\arraystretch}{1.1}
\resizebox{\textwidth}{!}{%
\begin{tabular}{llrrrrrrrr}
\toprule
\multirow{2}{*}{Method} & \multirow{2}{*}{Venue} & \multicolumn{4}{c}{WideIRSTD-Full} & \multicolumn{4}{c}{IRSTD-1k}\\
\cmidrule(lr){3-6}\cmidrule(lr){7-10}
 & & P$\uparrow$ & R$\uparrow$ & $F_1\uparrow$ & IoU$\uparrow$ & P$\uparrow$ & R$\uparrow$ & $F_1\uparrow$ & IoU$\uparrow$\\
\midrule
\multicolumn{10}{l}{\textbf{Open-Source MLLMs}} \\
InternVL3.5-38B & \raisebox{-0.16em}{\includegraphics[height=1.05em]{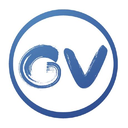}}\,'25 & 10.92 & 4.04 & 5.89 & 34.32 & 6.49 & 4.04 & 4.98 & 41.84 \\
Qwen3.5-9B & \raisebox{-0.16em}{\includegraphics[height=1.05em]{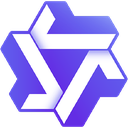}}\,'26 & 7.14 & 5.16 & 5.99 & 36.07 & 23.85 & 20.88 & 22.26 & 40.84 \\
Qwen3.5-27B & \raisebox{-0.16em}{\includegraphics[height=1.05em]{figures/model_icons/qwen.png}}\,'26 & 13.77 & 6.85 & 9.15 & 38.68 & 41.04 & 34.68 & 37.59 & 44.54 \\
Qwen3.8-27B & \raisebox{-0.16em}{\includegraphics[height=1.05em]{figures/model_icons/qwen.png}}\,'26 & 30.03 & 15.02 & 20.02 & 43.14 & 63.79 & 49.83 & 55.95 & 48.54 \\
LocateAnything-3B & \raisebox{-0.16em}{\includegraphics[height=1.05em]{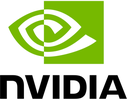}}\,'26 & 32.22 & 17.06 & 22.31 & 50.03 & \underline{79.02} & 38.05 & 51.36 & 54.40 \\
SenseNova-Vision-7B-MoT & \raisebox{-0.16em}{\includegraphics[height=1.05em]{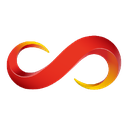}}\,'26 & 3.43 & 2.48 & 2.88 & 42.00 & 28.07 & 5.39 & 9.04 & 59.67 \\
\specialrule{0.8pt}{0pt}{0pt}
\multicolumn{10}{l}{\textbf{VLM-Based Grounding Models}} \\
Florence-2-large-ft & CVPR '24 & 6.90 & 3.66 & 4.78 & 64.39 & 11.43 & 14.81 & 12.90 & 68.22 \\
Grounding DINO-B & ECCV '24 & 9.33 & 5.50 & 6.92 & 66.85 & 10.93 & 11.45 & 11.18 & 69.85 \\
YOLO-World & CVPR '24 & 13.48 & 6.57 & 8.83 & \textbf{67.99} & 6.82 & \underline{75.08} & 12.51 & 69.65 \\
Rex-Omni & CVPR '26 & 16.04 & \textbf{39.06} & 22.74 & 67.01 & 48.37 & \underline{75.08} & 58.84 & \underline{70.34} \\
\midrule
\multicolumn{10}{l}{\textbf{Visual Reasoning and Active Search Methods}} \\
ZoomEye & EMNLP '25 & 19.01 & 34.74 & 24.58 & 65.56 & 48.40 & \textbf{81.48} & 60.73 & 69.07 \\
RSCoVLM-det & RS '26 & \underline{46.38} & 22.35 & \underline{30.17} & 63.42 & 77.70 & 72.73 & \underline{75.13} & 69.70 \\
ZoomEarth & CVPR '26 & 15.97 & 4.37 & 6.87 & 64.72 & 44.62 & 29.29 & 35.37 & 69.79 \\
DeepEyes & ICLR '26 & 4.34 & 4.30 & 4.32 & 62.28 & 26.90 & 28.62 & 27.73 & 69.01 \\
VisionReasoner & ICLR '26 & 18.78 & 21.98 & 20.25 & 66.63 & 37.73 & 54.88 & 44.72 & 68.33 \\
\midrule
\rowcolor{gray!8}
\textbf{IRSTD-Agent} & \textbf{Ours} & \textbf{71.63} & \underline{34.83} & \textbf{46.87} & \underline{67.07} & \textbf{81.09} & \underline{75.08} & \textbf{77.97} & \textbf{71.56} \\
\bottomrule
\end{tabular}}
\end{table*}

\paragraph{WideIRSTD-Full datasets.}
Direct full-image perception remains challenging on WideIRSTD-Full because tiny targets occupy only a small fraction of large infrared images and their weak responses can be lost during visual compression. As shown in Table~\ref{tab:dataset_comparison}, Qwen3.5-27B and Qwen3.8-27B achieve 9.15 and 20.02 F1-scores, while LocateAnything-3B reaches 22.31. Grounding models show a similar limitation; Rex-Omni attains 39.06 recall but only 22.74 F1-scores. In contrast, search-based methods benefit from acquiring local evidence. ZoomEye reaches 24.58 F1-scores, RSCoVLM-det reaches 30.17, and IRSTD-Agent achieves 46.87 F1-scores with 71.63 precision. Compared with direct Qwen3.5-27B inference using the same backbone, this is a 37.72 F1-scores gain, supporting the importance of active local observation in wide-area IRSTD.

\paragraph{IRSTD-1k datasets.}
On IRSTD-1k, the fixed $512\times512$ resolution substantially reduces the scale mismatch between the original image and the model input. Table~\ref{tab:dataset_comparison} shows that direct full-image models therefore become much stronger: Qwen3.8-27B reaches 55.95 F1-scores and Rex-Omni reaches 58.84. Active search remains beneficial, with ZoomEye reaching 60.73 F1-scores and RSCoVLM-det 75.13. IRSTD-Agent further achieves 81.09 precision, 75.08 recall, and 77.97 F1-scores. Compared with the WideIRSTD-Full results, this contrast suggests that active evidence acquisition is especially important when large image scale makes a single global observation insufficient.

\subsection{Qualitative analysis.}
\begin{figure}[!t]
\centering
\includegraphics[width=\linewidth]{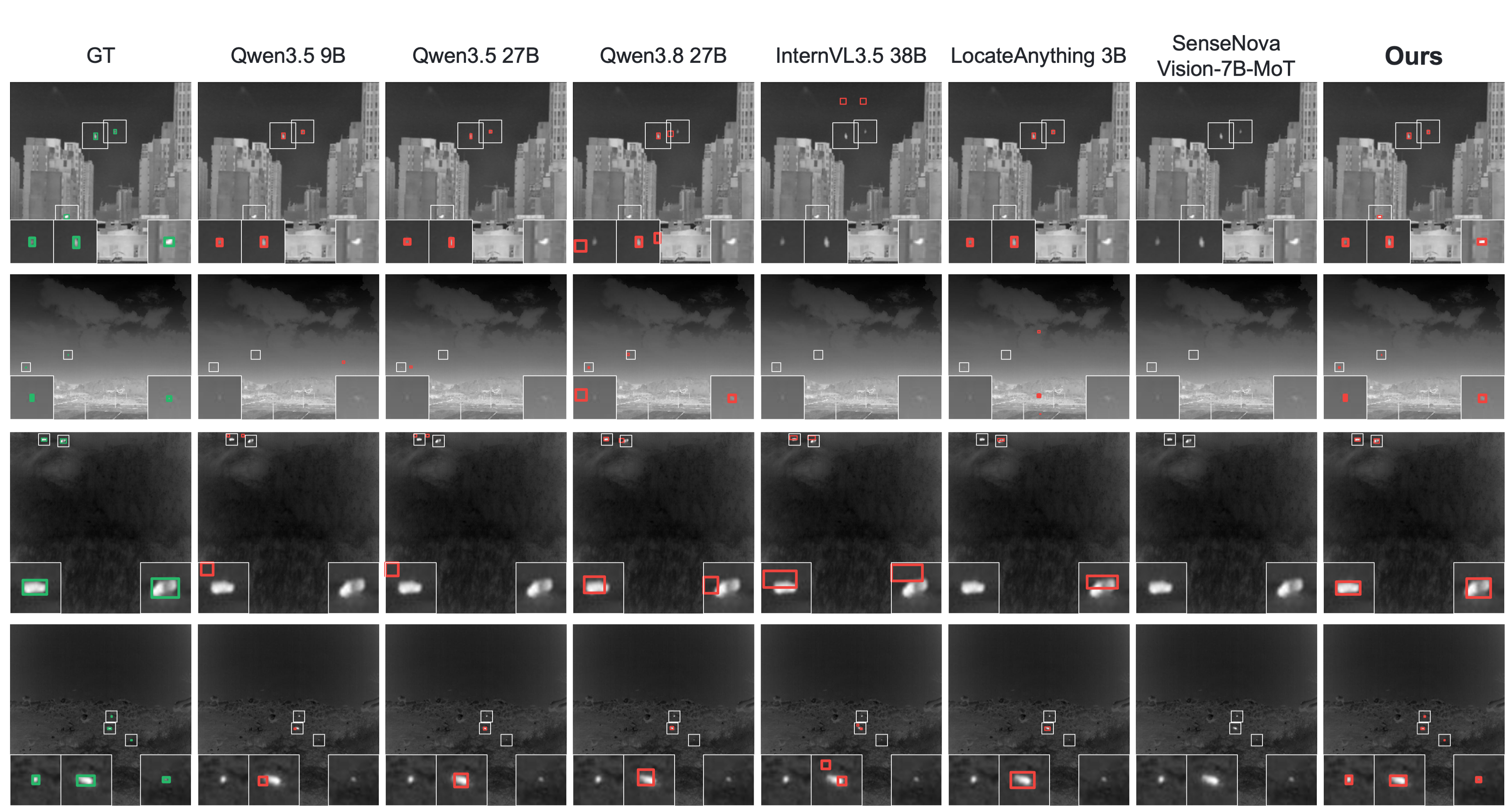}
\caption{Selected qualitative multi-target detection examples from WideIRSTD-Full (top two rows) and IRSTD-1k (bottom two rows) datasets. Columns compare ground truth, six full-image VLM baselines, and our method. \textcolor[rgb]{0.0,0.35,0.0}{Green boxes} indicate ground truth and \textcolor{red}{red boxes} indicate predictions. White boxes mark regions enlarged in the insets along the bottom of each image.}
\label{fig:qualitative_multitarget}
\end{figure}

Figure~\ref{fig:qualitative_multitarget} presents representative multi-target detection results on WideIRSTD-Full (top two rows) and IRSTD-1k (bottom two rows). The white regions highlight target-containing areas and are enlarged in the bottom insets, where green boxes denote ground-truth targets and red boxes denote model predictions. Direct full-image MLLMs frequently miss faint targets or produce spatially offset detections. In contrast, IRSTD-Agent recovers more weak targets through local visual search and yields more accurate localization in these challenging regions. Nevertheless, the examples also show that precise localization remains difficult for extremely small and visually ambiguous targets.

\section{Conclusion}
In this work, we investigate the challenges of infrared small target detection with multimodal large language models and highlight the importance of actively acquiring fine-grained visual evidence for precise localization. We propose IRSTD-Agent, an agentic framework that formulates detection as a dynamic visual search process. Five complementary visual tools enable candidate discovery, adaptive observation, target localization, hypothesis rejection, and target extent refinement. We further introduce Zoom-guided Interaction Learning, which uses annotation-derived interaction trajectories to teach the model to select appropriate tools and their arguments. Extensive experiments on WideIRSTD-Full and IRSTD-1k demonstrate the effectiveness of our framework, demonstrating detection performance superior to other state-of-the-art MLLM-based methods.

\textbf{Limitations and Future Work.} Sequential tool interactions introduce additional inference overhead, and extremely small or visually ambiguous targets in IRSTD task remain challenging. Future work could investigate more efficient search policies and adaptive computation allocation to improve the balance between detection accuracy and computational cost. 

\label{sec:main_end}

\bibliography{iclr2027_conference}
\bibliographystyle{iclr2027_conference}

\clearpage
\appendix
\section*{Appendix}

\section{Implementation Details}
\label{app:implementation_details}

\subsection{Tools and Coordinate Interface}
For implementation, we represent the executed view by its boundary coordinates $C=[u_1,v_1,u_2,v_2]$, with width $w_C=u_2-u_1>0$ and height $h_C=v_2-v_1>0$. This is equivalent to the center-and-size view representation used in Section~\ref{sec:agent_design} after boundary handling. Crop-normalized and original-image coordinates are related by
\begin{equation}
 T_C(x,y)=(u_1+xw_C,v_1+yh_C),\qquad
 T_C^{-1}(X,Y)=\left(\frac{X-u_1}{w_C},\frac{Y-v_1}{h_C}\right).
\end{equation}
A requested zoom size denotes the side length of the next view in original-image pixels. Each observation is extracted directly from the original image after rounding and boundary handling, rather than recursively from a previously resized crop.

In the main joint-policy implementation, \textsc{Proposal} is invoked as a policy-controlled tool, after which \textsc{Zoom} selects a candidate-relative location and crop size for the next observation. Some controlled experiments use an earlier fixed-queue configuration in which proposals are initialized before search. In both cases, tool execution updates the current view and branch-local interaction history while preserving accumulated image-level detections. \textsc{Detect} and \textsc{Drop} terminate the current branch, with \textsc{Drop} producing no detection. \textsc{Detect} outputs normalized target centers $\{(x_j,y_j)\}$ in the current crop, which are mapped to original-image coordinates by $T_C$ before refinement.

\subsection{Proposal Tool}
Given a grayscale native-resolution view $J$, we first normalize its intensity range to obtain $\bar J$. The spatial branch combines four complementary responses:
\begin{equation}
R_{\rm sp}
=
\frac{1}{4}\Big[
\mathcal N(R_{\rm TH})
+\mathcal N(R_{\rm DoG})
+\mathcal N(R_{\rm LC})
+\mathcal N(R_{\rm PC})
\Big],
\end{equation}
where $R_{\rm TH}$, $R_{\rm DoG}$, $R_{\rm LC}$, and $R_{\rm PC}$ denote multi-scale top-hat, difference-of-Gaussians, local standardized contrast, and patch-contrast responses, respectively, and $\mathcal N(\cdot)$ denotes response normalization.

For the frequency branch, we use spectral-residual saliency. Let
$A=|\mathcal F(\bar J)|$ and $\phi=\arg\mathcal F(\bar J)$. The response is
\begin{equation}
R_{\rm fft}
=
\mathcal N\!\left(
G *
\left|
\mathcal F^{-1}
\left(
e^{\,L-\bar L+\mathrm i\phi}
\right)
\right|^2
\right),
\qquad
L=\log(A+\epsilon),
\end{equation}
where $\bar L$ is a locally smoothed log-amplitude spectrum and $G$ denotes spatial smoothing.

The two branches are fused as
\begin{equation}
R
=
\lambda R_{\rm sp}
+
(1-\lambda)R_{\rm fft}.
\end{equation}
Local peaks in $R$ are then ranked and spatially suppressed to produce the proposal set $\mathcal Q_t$. All proposal operations are image-based and contain no learned parameters.

\subsection{Refinement Tool and Output Assembly}
Given a detected center $p_i$, \textsc{Refine} extracts a local neighborhood from the original image and estimates its background appearance. Let $\hat B_i(p)$ denote the estimated local background and define the residual response as
\begin{equation}
    r_i(p)=I(p)-\hat B_i(p).
\end{equation}
An adaptive threshold $\theta_i$, estimated from the local residual statistics, is then used to obtain target-supporting pixels. Starting from the response nearest to $p_i$, we retain the connected component
\begin{equation}
    \Omega_i
    =
    \operatorname{CC}_{p_i}
    \left(\{p:r_i(p)\geq\theta_i\}\right),
\end{equation}
where $\operatorname{CC}_{p_i}$ denotes the connected region associated with the detected target center.

The spatial extent of $\Omega_i$ defines the refined bounding box,
\begin{equation}
    \tilde b_i=\operatorname{BBox}(\Omega_i),
    \qquad
    M_i(p)=\mathbf 1[p\in\Omega_i],
\end{equation}
where $M_i$ is the corresponding foreground mask. If no valid connected region is obtained, the initialized local box around $p_i$ is retained.

For multiple detections, the image-level segmentation map is formed by
\begin{equation}
    M(p)=\bigvee_i M_i(p).
\end{equation}
Bounding-box predictions are subsequently merged using NMS, while mask assembly preserves the image-derived foreground regions rather than filling the refined boxes.

\subsection{Zoom-Guided interaction learning}
Following Equations~\ref{eq:trajectory}--\ref{eq:sft_dataset}, each retained trajectory $\tau_i$ is constructed from a source image $I_i$ and its training annotations:
\begin{equation}
 \tau_i=\{(I_i[C_{i,t}],m_{i,t},a_{i,t}^\star)\}_{t=0}^{T_i},
 \qquad
 \mathcal D_{\mathrm{SFT}}=\{\tau_i\}_{i=1}^{N}.
\end{equation}
Here $i$ indexes trajectories rather than distinct images, since multiple trajectories may be constructed from the same source image.

Connected components of the training mask define target boxes $g_j$ and centers $p_j^\star$. For a view $C$, we distinguish fully visible targets from those that only intersect the view:
\begin{equation}
 V(C)=\{j:g_j\subseteq C\},\qquad
 U(C)=\{j:\operatorname{area}(g_j\cap C)>0\}.
\end{equation}

Positive trajectories consist of target-preserving \textsc{Zoom} transitions that progressively acquire more informative local observations. Offset-recovery examples deliberately displace the initial candidate from the target center, requiring localization from visual evidence rather than direct proposal copying. At a terminal positive view, \textsc{Detect} supervises the normalized centers of all fully visible targets:
\begin{equation}
 a_t^\star=\mathtt{DETECT}
 \left(\{T_{C_t}^{-1}(p_j^\star):j\in V(C_t)\}\right).
\end{equation}

Negative trajectories are constructed from candidate branches containing no annotated target. A \textsc{Drop} label is assigned only when $U(C_t)=\varnothing$, ensuring that partially visible targets are not treated as negatives. Multi-target views retain all valid targets in the corresponding \textsc{Detect} supervision. \textsc{Proposal} and \textsc{Zoom} actions are included along the trajectory according to the executed search process.

Each structured action is serialized into a token sequence
$y_{i,t}^\star$. The token-level SFT objective is
\begin{equation}
 \mathcal L_{\mathrm{SFT}}
 =-\sum_{i=1}^{N}\sum_{t=0}^{T_i}
 \sum_{\ell=1}^{|y_{i,t}^\star|}
 \log \pi_\theta
 \left(
 y_{i,t,\ell}^\star
 \mid I_i[C_{i,t}],m_{i,t},y_{i,t,<\ell}^\star
 \right).
\end{equation}
Prompt tokens are masked from the loss. Separate trajectory sets and
checkpoints are constructed for each dataset.
\subsection{Baseline Adaptation}

We follow the method groups in Table~\ref{tab:dataset_comparison} and adapt each baseline to infrared small-target detection while preserving its original inference paradigm whenever possible.

\paragraph{MLLM and grounding baselines.}
InternVL3.5, Qwen3.5/3.8, LocateAnything, and SenseNova-Vision use released checkpoints without infrared-specific fine-tuning. They directly process the full image and are prompted to return all visible infrared targets as bounding boxes, with model-specific coordinate formats mapped back to the original image.

Florence-2, Grounding DINO, YOLO-World, and Rex-Omni likewise use released weights and perform full-image multi-target localization with an infrared-target query. Their native task interfaces and decoding procedures are retained wherever applicable.

\paragraph{Visual reasoning and active search baselines.}
ZoomEye, ZoomEarth, DeepEyes, and VisionReasoner retain their original search, cropping, or reasoning procedures, while their task interfaces are adapted to produce infrared-target detections. ZoomEye uses Qwen3.5-27B as its underlying multimodal model and performs hierarchical visual search before localization. Predictions produced in local views are mapped back to the original image coordinates for evaluation.

RSCoVLM-det is the only baseline in this group that receives dataset-specific training. Separate LoRA adapters are trained for WideIRSTD-Full and IRSTD-1k using annotations converted from the corresponding training masks.

\paragraph{Specialized IRSTD detectors.}
The specialized detectors reported in Appendix~\ref{sec:specialized_comparison}, including RDIAN, LW-IRSTNet, DNANet, MSHNet, and NSFPN, are trained separately on each dataset using their original supervision. Their full-image and sliding-window evaluation protocols are reported separately in the corresponding appendix section.

\section{More Analysis}
\label{app:comparisons}

\subsection{Comparison with Specialized IRSTD Methods}
\label{sec:specialized_comparison}

Table~\ref{tab:specialized_wide} compares specialized IRSTD detectors with IRSTD-Agent on the full WideIRSTD-Full test set and its large-image subset. The full set contains 2,000 images and 5,328 targets. The Large subset contains 44 images with original area $A=W\times H>1024^2$, comprising 99 targets. For these 44 images, we report both the original full-image resize protocol and a native-resolution $256\times256$ sliding-window alternative for the specialized detectors.
\begin{table}[h!]
\centering
\caption{Specialized IRSTD methods and IRSTD-Agent on WideIRSTD-Full. Full and Large (original) use full-image resize, while Large (sliding) uses native-resolution $256\times256$ sliding windows for the specialized detectors. The two Large settings contain the same 44 images and 99 targets. IRSTD-Agent is unchanged in both columns. Values are percentages; IoU is averaged over matched targets only.}
\label{tab:specialized_wide}
\scriptsize
\setlength{\tabcolsep}{2.4pt}
\renewcommand{\arraystretch}{1.12}
\resizebox{\textwidth}{!}{%
\begin{tabular}{@{}lcccccccccccc@{}}
\toprule
Method & \multicolumn{4}{c}{Full (2,000)} & \multicolumn{4}{c}{Large: original (44)} & \multicolumn{4}{c}{Large: sliding $256$ (44)}\\
\cmidrule(lr){2-5}\cmidrule(lr){6-9}\cmidrule(lr){10-13}
& P & R & $F_1$ & IoU & P & R & $F_1$ & IoU & P & R & $F_1$ & IoU\\
\midrule
RDIAN & 72.66 & 46.43 & 56.66 & 60.25 & 33.33 & 1.01 & 1.96 & 14.06 & 48.60 & 52.53 & 50.49 & 69.54\\
U-Net & 78.35 & 49.32 & 60.54 & 60.27 & 20.00 & 1.01 & 1.92 & 14.06 & 54.93 & 39.39 & 45.88 & 75.39\\
ResUNet & 77.44 & 50.06 & 60.81 & 59.80 & 20.00 & 1.01 & 1.92 & 8.33 & 58.67 & 44.44 & 50.57 & 65.99\\
AttU-Net & 78.66 & 48.16 & 59.74 & 59.40 & 12.50 & 1.01 & 1.87 & 25.00 & 55.56 & 45.45 & 50.00 & 66.89\\

MLCL-Net & 78.69 & 41.85 & 54.64 & 59.39 & 20.00 & 1.01 & 1.92 & 25.00 & 44.92 & 53.54 & 48.85 & 64.13\\

\midrule
\rowcolor{gray!8}
\textbf{IRSTD-Agent} & 71.63 & 34.83 & 46.87 & 67.07 & 78.26 & 54.55 & 64.29 & 75.34 & 78.26 & 54.55 & 64.29 & 75.34\\
\bottomrule
\end{tabular}}
\end{table}

For Full and Large (original), the specialized detectors follow the same full-image $256\times256$ resize protocol used in their standard evaluation, with predictions mapped back to the original image. For Large (sliding), the same trained checkpoints are evaluated without retraining using native-resolution $256\times256$ windows with stride 128. Predictions from overlapping windows are merged on the original image. IRSTD-Agent retains its original active local-observation pipeline, and its Large result is repeated for reference.

Specialized IRSTD detectors are highly task-specific and perform strongly under their conventional input setting. On the full test set, their F1 scores range from 54.64 to 60.81, exceeding the 46.87 F1 of IRSTD-Agent. However, this advantage does not transfer directly to large images. Under full-image resize, all specialized detectors recover only 1.01\% of the targets in the Large subset, with F1 around 1.9, showing that aggressive resizing can largely remove the already weak small-target evidence.

Native-resolution sliding-window inference substantially recovers their performance, increasing F1 to 45.88--50.57 and recall to 39.39--53.54. Nevertheless, a considerable fraction of targets remains missed, and all specialized detectors remain below IRSTD-Agent on this subset, which reaches 54.55 recall and 64.29 F1. These results highlight the strong dependence of specialized IRSTD models on the observation scale: preserving local resolution is essential for large-image detection, while a fixed sliding-window strategy still provides less flexible evidence acquisition than the active search used by IRSTD-Agent.

\subsection{Segmentation Results}
\label{sec:segmentation_results}

Table~\ref{tab:segmentation_wide} reports segmentation performance on WideIRSTD-Full (2,000 images) and IRSTD-1k (201 images). For the MLLM, grounding, visual-search, and IRSTD-Agent pipelines, detected target centers are converted to foreground masks using the same center-to-mask refinement procedure. Therefore, the full-image MLLM results in this table include mask generation beyond their native-box detection outputs reported in the main text.

We report image-averaged mask IoU (nIoU), foreground Dice, component-level detection probability ($P_d$), and false-alarm rate ($F_a$). nIoU is averaged over images with a nonempty mask union, while Dice is computed over foreground pixels across the test set. $P_d$ uses one-to-one component-centroid matching within three pixels. $F_a$ measures foreground pixels belonging to unmatched predicted components relative to the total number of image pixels and is reported as $10^6F_a$; lower values indicate fewer false alarms.

\par\medskip\noindent\begin{minipage}{\textwidth}
\makeatletter\def\@captype{table}\makeatother
\centering
\caption{Segmentation performance on WideIRSTD-Full and IRSTD-1k. nIoU, Dice, and $P_d$ are reported as percentages, while $F_a$ is reported as $10^6F_a$. All methods use the same center-to-mask refinement for segmentation evaluation.}
\label{tab:segmentation_wide}
\small
\setlength{\tabcolsep}{3pt}
\renewcommand{\arraystretch}{1.08}
\resizebox{\textwidth}{!}{%
\begin{tabular}{lrrrrrrrr}
\toprule
Method & \multicolumn{4}{c}{WideIRSTD-Full (2,000)} & \multicolumn{4}{c}{IRSTD-1k (201)}\\
\cmidrule(lr){2-5}\cmidrule(lr){6-9}
 & nIoU$\uparrow$ & Dice$\uparrow$ & $P_d\uparrow$ & $F_a\downarrow$ & nIoU$\uparrow$ & Dice$\uparrow$ & $P_d\uparrow$ & $F_a\downarrow$\\
\midrule
\multicolumn{9}{l}{\textbf{Open-Source MLLMs}} \\
InternVL3.5-38B & 12.50 & 20.95 & 13.59 & 8.91 & 20.49 & 32.88 & 35.35 & 7.38 \\
Qwen3.5-9B & 12.30 & 21.35 & 14.88 & 16.86 & 37.83 & 42.35 & 52.53 & 17.37 \\
Qwen3.5-27B & 19.06 & 31.17 & 22.07 & 11.31 & 42.71 & 45.47 & 58.92 & 9.62 \\
Qwen3.8-27B & 19.07 & 32.90 & 23.25 & 12.69 & 48.97 & 57.22 & 68.35 & 9.57 \\
LocateAnything-3B & 15.78 & 23.42 & 17.83 & 11.63 & 27.48 & 35.94 & 37.37 & 3.28 \\
SenseNova-Vision-7B-MoT & 2.45 & 4.96 & 2.85 & 12.72 & 4.62 & 6.75 & 5.39 & 5.67 \\
\midrule
\multicolumn{9}{l}{\textbf{VLM-Based Grounding Models}} \\
Florence-2-large-ft & 4.09 & 9.00 & 3.55 & 10.37 & 9.39 & 18.11 & 13.80 & 48.89 \\
Grounding DINO-B & 7.35 & 11.04 & 5.29 & 14.01 & 8.48 & 17.48 & 10.44 & 30.90 \\
YOLO-World & 4.92 & 13.24 & 6.55 & 19.18 & 25.62 & 30.78 & 73.06 & 379.23 \\
Rex-Omni & 28.58 & 34.32 & 38.95 & 67.64 & 49.09 & 54.52 & 74.07 & 53.52 \\
\midrule
\multicolumn{9}{l}{\textbf{Visual Reasoning and Active Search}} \\
ZoomEye & 25.14 & 34.47 & 34.98 & 48.98 & 54.36 & 57.81 & 80.13 & 41.54 \\
RSCoVLM-det & 26.34 & 29.34 & 22.22 & 12.98 & 51.93 & 57.46 & 70.71 & 12.41 \\
ZoomEarth & 4.03 & 8.78 & 4.43 & 5.74 & 20.38 & 33.71 & 28.96 & 16.06 \\
DeepEyes & 4.07 & 6.91 & 4.34 & 19.11 & 17.53 & 28.88 & 27.95 & 28.37 \\
VisionReasoner & 20.07 & 27.40 & 22.03 & 30.24 & 35.14 & 43.39 & 54.21 & 58.44 \\
\midrule
\rowcolor{gray!8}
\textbf{IRSTD-Agent} & 31.71 & 37.67 & 34.68 & 13.35 & 52.19 & 57.90 & 76.09 & 17.61 \\
\bottomrule
\end{tabular}}
\end{minipage}\par\medskip

On WideIRSTD-Full, IRSTD-Agent achieves the highest nIoU (31.71) and Dice (37.67), indicating stronger foreground overlap under the large-image setting. Rex-Omni attains a higher $P_d$ of 38.95, but with a substantially larger false-alarm rate (67.64 versus 13.35), suggesting that its higher target recovery is accompanied by more spurious foreground responses.

On IRSTD-1k, IRSTD-Agent reaches 52.19 nIoU, 57.90 Dice, 76.09 $P_d$, and 17.61 $F_a$. ZoomEye achieves slightly higher nIoU and $P_d$ (54.36 and 80.13), but also produces considerably more false alarms ($F_a=41.54$). Overall, the results show that segmentation quality depends not only on target recovery, but also on the spatial accuracy and false foreground introduced by the complete detection-and-refinement pipeline.

\section{Prompt Templates}
\label{app:prompt_template}

\subsection{IRSTD-Agent}
The system and observation prompts below use the center-output interface. Line breaks are adjusted for layout. The proposal instruction is appended to the original system prompt in compact form.

\paragraph{System prompt.}
{\small
\begin{verbatim}
You are an infrared-small-target gaze agent. Inspect only the
supplied crop. Coordinates are normalized to the CURRENT crop.
Use ZOOM when additional local evidence is needed. Use DETECT when
the visible targets can be reliably localized, and DROP when the
current hypothesis is unsupported. Emit exactly one JSON action
and no prose:
ZOOM {type,x,y,size}; DETECT {type,targets:[{x,y}]};
DROP {type}; or PROPOSAL {type}.
PROPOSAL runs RF/FFT proposals on the current crop when
proposal_allowed=true and additional candidate hypotheses are
needed. Proposal points guide search; they are not detections.
\end{verbatim}
}

\paragraph{Observation prompt.}
Each user message contains the current crop and the following text. Placeholders are filled with runtime values; \texttt{history} contains parsed actions from the current branch.
{\small
\begin{verbatim}
image_size={image_size}; crop_original={crop_original};
crop_size={crop_size}; step={step}; history={history}.
Choose the next action.
\end{verbatim}
}
Here \texttt{image\_size} is $[W,H]$, \texttt{crop\_original} is $[u_1,v_1,u_2,v_2]$, \texttt{crop\_size} is $[u_2-u_1,v_2-v_1]$, and \texttt{step} is the zero-based branch depth. When an anchor is active, the observation appends:
{\small
\begin{verbatim}
proposal_anchor_original={anchor}; Follow the current proposal
anchor when choosing which local hypothesis to inspect; do not
switch to a distant bright target belonging to another proposal.
\end{verbatim}
}
The anchor uses original-image coordinates. The observation also includes \texttt{proposal\_allowed} and, after a local \textsc{Proposal} call, \texttt{local\_proposals\_original}, a list of original-image candidate centers.

\paragraph{Detection action instruction.}
When \textsc{Detect} is selected, IRSTD-Agent localizes all visible targets in the current crop. The detection-specific instruction is summarized below:
{\small
\begin{verbatim}
Detect every fully visible infrared small target in this crop.
Return all target centers normalized to the CURRENT crop in [0,1],
not box corners or original-image coordinates.
Output one JSON action:
{"type":"DETECT","targets":[{"x":...,"y":...}]}
If the crop is background, output {"type":"DROP"}.
\end{verbatim}
}
IRSTD-Agent predicts the target centers, while \textsc{Proposal} only supplies candidate locations. The detection instruction belongs to the same policy prompt. The executor maps the predicted centers to the original image before subsequent localization refinement.

\paragraph{Structured responses.}
Illustrative action formats are:
{\small
\begin{verbatim}
{"type":"PROPOSAL"}
{"type":"ZOOM","x":0.5,"y":0.5,"size":64}
{"type":"DETECT","targets":[{"x":0.3,"y":0.4},
                            {"x":0.7,"y":0.6}]}
{"type":"DROP"}
\end{verbatim}
}
Zoom coordinates are crop-normalized; \texttt{size} is the requested side length in original-image pixels. \textsc{Detect} and \textsc{Drop} end the current branch, after which the executor advances to the next candidate.

\subsection{Full-Image MLLM and Grounding Prompts}
\paragraph{Full-image MLLM prompt.}
The task instruction requests all infrared targets from the full image. A compact rendering is:
{\small
\begin{verbatim}
Detect all compact infrared small targets, including dim targets.
Exclude broad bright regions, edges, and background texture.
Return one tight bounding box per target in the required
coordinate format, and an empty list if no target is visible.
Output only the requested structured result.
\end{verbatim}
}
Model-specific output conventions are retained rather than imposing one literal prompt on all models. For example, the Qwen native-coordinate interface requests JSON boxes with each axis independently normalized to $[0,1000]$ and supplies the image dimensions; parsed boxes are mapped back to original-image coordinates.

\paragraph{Grounding prompt.}
These methods receive an infrared-target text query through their native grounding interfaces rather than the agent's action prompt. The category query used by YOLO-World and Rex-Omni is:
{\small
\begin{verbatim}
infrared small target
\end{verbatim}
}
Florence-2 uses its \texttt{<CAPTION\_TO\_PHRASE\_GROUNDING>} task with an infrared-target query, while Grounding DINO receives an infrared-target caption. Their native task formatting and output parsers are preserved.

\section{Large Language Model Usage}
\label{app:ai_assistance}
AI tools were used for language polishing and assistance with LaTeX formatting, including figure and table layout. Portions of the code were implemented and modified with assistance from GPT-5 and GPT-6 Codex. The authors take full responsibility for the final manuscript, code, and all reported results.

\end{document}